\documentclass[pdflatex,sn-mathphys-num]{sn-jnl}

\usepackage{graphicx}%
\usepackage{multirow}%
\usepackage{amsmath,amssymb,amsfonts}%
\usepackage{amsthm}%
\usepackage{mathrsfs}%
\usepackage[title]{appendix}%
\usepackage{xcolor}%
\usepackage{textcomp}%
\usepackage{manyfoot}%
\usepackage{booktabs}%
\usepackage{algorithm}%
\usepackage{algorithmicx}%
\usepackage{algpseudocode}%
\usepackage{listings}%
\usepackage{tabularx}
\usepackage{multirow}
\usepackage{makecell}
\usepackage{float}
\usepackage{fancyvrb}
\usepackage{tcolorbox}
\usepackage{siunitx}    
\usepackage{xcolor}     
\usepackage{fvextra}  
\usepackage{colortbl}   
\usepackage{caption}    
\usepackage{soul}
\sethlcolor{yellow}  
\tcbuselibrary{listings, breakable}

\theoremstyle{thmstyleone}%
\theoremstyle{thmstyletwo}%

\theoremstyle{thmstylethree}%

\begin{document}

\title[Article Title]{DialogPII: A multilingual dataset of synthetic dialog transcripts to detect personal information}


\author*[1]{\fnm{Roland} \sur{Roller}}\email{roland.roller@dfki.de}
\author[1,2]{\fnm{Vera} \sur{Czehmann}}
\author[2]{\fnm{Derya} \sur{Erman}}
\author[3]{\fnm{Luke} \sur{Flanagan}}
\author[1,2]{\fnm{Ibrahim} \sur{Baroud}}
\author[4]{\fnm{Frédéric} \sur{Blain}}
\author[5]{\fnm{Viviana} \sur{Cotik}}
\author[6]{\fnm{Eletta} \sur{Giusto}}
\author[1]{\fnm{Akhil} \sur{Juneja}}
\author[7]{\fnm{Mariana} \sur{Neves}}
\author[2]{\fnm{Maria} \sur{Słowińska}}
\author[1]{\fnm{Christine} \sur{Hovhannisyan}}
\author[1]{\fnm{Aaron Louis} \sur{Eidt}}
\author[1,2]{\fnm{Lisa} \sur{Raithel}}
\author[1,2]{\fnm{Sebastian} \sur{Möller}}
\author*[3]{\fnm{Maija} \sur{Poikela}}\email{maija.poikela@bih-charite.de}

\affil[1]{\orgname{German Research Center for Artificial Intelligence (DFKI)}, \state{Berlin}, \country{Germany}}
\affil[2]{\orgname{Technical University Berlin}, \orgaddress{\city{Berlin}, \country{Germany}}}
\affil[3]{\orgname{Berlin Institute of Health (BIH)}, \orgaddress{\city{Berlin}, \country{Germany}}}
\affil[4]{\orgname{Tilburg University}, \orgaddress{\city{Tilburg}, \country{Netherlands}}}
\affil[5]{\orgname{Universidad de Buenos Aires}, \orgaddress{\city{Buenos Aires}, \country{Argentina}}}
\affil[6]{\orgname{Independent Researcher}, \orgaddress{\city{Berlin}, \country{Germany}}}
\affil[7]{\orgname{Bundesinstitut für Risikobewertung (BfR)}, \orgaddress{\city{Berlin}, \country{Germany}}}


\abstract{Conversational data collected in domains such as healthcare or social sciences is a valuable resource for research and automated analysis. However, responsible data sharing requires the detection and removal of personally identifiable and sensitive information to protect individual privacy. To support the development and evaluation of automatic de-identification systems, we present DialogPII, a multilingual dataset of synthetic dialogs and speech-derived transcripts for personal information detection. DialogPII covers eight interaction scenarios (emergency calls, medical anamnesis interviews, therapy sessions, insurance communication, customer support, clinical interviews regarding an AI-supported dashboard, police reports, and group therapy discussions), 19 entity types, and 11 languages (English, Arabic, Finnish, French, German, Hindi, Italian, Polish, Portuguese, Spanish, and Turkish). Dialogs were generated semi-automatically using large language models, manually curated for plausibility and diversity, and localized to country- and city-specific contexts. All dialogs were additionally converted to speech via text-to-speech synthesis, transcribed with Whisper, and annotated through automatic projection and manual correction, yielding aligned written and speech-derived resources across all languages. We further release baseline multilingual named entity recognition models and provide technical validation through inter-annotator agreement analysis, translation quality evaluation, annotation projection assessment, and benchmark experiments with transformer-based sequence labeling models.}




\maketitle

\section{Background \& Summary}\label{sec1}

Conversational data is increasingly collected and analyzed in domains such as healthcare, social sciences, and customer support. Interviews, therapy sessions, emergency calls, medical consultations, and online support groups provide valuable information for qualitative analysis, documentation, and machine learning applications. However, these interactions frequently contain personally identifiable information (PII), including names, locations, professions, identification numbers, medical conditions, and social relationships. Without careful removal of sensitive information, such data cannot be responsibly shared or reused for scientific purposes. Reliable de-identification tools can support the process of unlocking its research potential, particularly when audio recordings or detailed transcripts are involved. 

Existing research on text anonymization and de-identification has primarily focused on medical documents such as clinical notes or discharge summaries \cite{uzuner2007evaluating} \cite{stubbs2015annotating} \cite{dernoncourt2017identification} \cite{richter2023distributable} \cite{lohr2024identifying}. Widely used annotation standards, including personal health identifiers (PHI) as defined under HIPAA [45 CFR §164.514], mainly target formal medical documentation and are less suited for conversational language. While written documents bring their own challenges, spoken interactions add complexity through fragmented sentences, informal language, interruptions, speaker changes, contextual references, and the spontaneous, unguarded disclosure of sensitive information. These aspects make direct transfer from document-oriented de-identification frameworks to conversational settings non-trivial. 

The detection of sensitive information in dialog transcripts has received comparatively less attention. Cohn et al. \cite{cohn2019audio} introduced the task of audio de-identification, formulating it as a named entity recognition (NER) task on transcripts produced by automatic speech recognition (ASR) systems. Their benchmark pipeline was evaluated on the Switchboard \cite{godfrey1992switchboard} and Fisher \cite{cieri2004fisher} corpora using PHI-oriented labels extended to include categories such as profession, age, and organization. Subsequent work has explored related settings: NER-based anonymization for French audio \cite{baril2022named}, call-center transcript anonymization \cite{kaplan2020may} \cite{gouvea2023trustera}, and content-level rewriting of long-form telephone conversations to suppress re-identification risks beyond voice \cite{aggazzotti2026content}. Work on de-identification of call transcripts for LLM fine-tuning \cite{gardiner2024data} and Indonesian speech de-identification \cite{abdjul2025indonesian} further demonstrates the growing interest in this direction. Despite this progress, existing resources remain largely monolingual, English-centric, and limited in both entity schema breadth and dialog scenario coverage. 

Multilingual resources for conversational de-identification remain particularly limited. General-purpose multilingual NER benchmarks such as MultiCoNER \cite{malmasi2022semeval} and MultiCoNER 2 \cite{fetahu2023semeval} cover a range of languages and entity types but target general web and search-query text, without addressing sensitive information or the dialog domain. The multilingual medical anonymization benchmark MultiGraSCCo \cite{baroud2026multigrascco}, more closely related to our work, covers direct and indirect personal identifiers across ten languages using a translation-based methodology -- confirming that high-quality multilingual anonymization data can be created and transferred synthetically -- but focuses on written medical records rather than spoken or conversational data. Recent work on multilingual PII annotation \cite{meena2025scalable} spans a broad range of locales but similarly targets general user-generated text rather than conversational transcripts. 

The scarcity of publicly available conversational datasets is partly a consequence of the problem we try to address: real-world dialogs require ethical approval, informed consent, and extensive data protection procedures before they can be shared. Anonymizing interview data is itself challenging, particularly in sensitive contexts where detailed personal narratives are central to the research value of the material \cite{saunders2015anonymising}. Synthetic data generation provides a promising alternative, enabling the creation of reproducible and shareable resources while sidestepping disclosure of real personal information \cite{hiebel2023can}. Crucially, synthetic datasets do not merely substitute for real data under legal constraint, but they allow researchers to study and develop anonymization systems that, once deployed, make sharing of genuine conversational data feasible in the first place. 

To fill this gap, we present a multilingual dataset of synthetic dialogs and speech-derived transcripts for personal information detection in conversational data. The dataset supports research on multilingual de-identification, conversational NER, dialog anonymization, and privacy-preserving NLP. Dialogs are generated semi-automatically using large language models and manually curated to improve plausibility, linguistic quality, and diversity. To reduce repetitive patterns commonly observed in synthetic conversations \cite{long-etal-2024-llms}, dialogs are generated across multiple scenarios reflecting different interaction types and disclosure situations: emergency calls, police reports, insurance communication, therapy sessions, medical anamnesis interviews, customer support interactions, and interviews related to clinical AI systems. The dataset additionally includes extended multi-speaker group therapy conversations. Compared to prior work on dialog de-identification \cite{cohn2019audio} \cite{kaplan2020may}, our dataset is substantially broader in language coverage, scenario diversity, and entity schema, and uniquely provides aligned written and speech-derived transcripts across 11 languages. 

\begin{figure*}[t]
    \centering
    \includegraphics[width=1.0\linewidth]{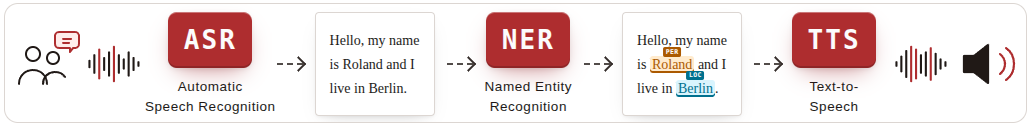}
    \caption{Overview of an exemplary dialog processing pipeline: From a spoken conversation, converted to text (transcript) using automatic speech recognition (ASR) and the detection of sensitive content via (our) named entity recognition (NER). After the removal of sensitive information, the dialog could technically be converted back into speech with the help of a Text-to-Speech (TTS) engine, or using the original audio, removing sensitive segments at particular time stamps. }
    \label{fig:asr_pipeline}
\end{figure*}

The dataset contains dialogs and aligned speech-derived transcripts in English, Arabic, Finnish, French, German, Hindi, Italian, Polish, Portuguese, Spanish, and Turkish. Dialogs were localized to country- and city-specific contexts to increase linguistic and cultural variation, similarly as in MultiGraSCCO \cite{baroud2026multigrascco}. Personal information is annotated using a dialog-oriented schema that combines concepts from existing de-identification frameworks with additional categories relevant to conversational settings, including professions, social relationships, and product references, covering 19 entity types in total. To support research on dialog anonymization pipelines, conversations are additionally converted to synthetic speech via text-to-speech synthesis and transcribed using automatic speech recognition. Personal information annotations are projected onto the generated transcripts and manually curated, yielding aligned resources across written dialogs, speech-derived transcripts, and multilingual annotations. 

In addition to the dataset, we release baseline multilingual NER models trained on the curated annotations. Technical validation includes inter-annotator agreement analysis, evaluation of translation quality, assessment of annotation projection accuracy, and multilingual benchmark experiments across all languages. The dataset is intended to facilitate future research on multilingual de-identification, conversational privacy protection, speech processing, and robust sensitive information detection in conversational AI systems, as depicted in Figure \ref{fig:asr_pipeline}. 

\section{Methods}\label{sec2}

\begin{figure*}[t]
    \centering
    \includegraphics[width=1.0\linewidth]{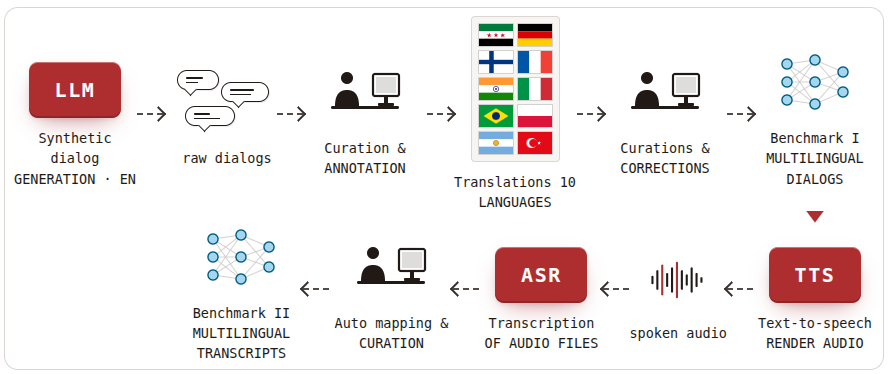}
    \caption{Overview of the complete pipeline from data generation, curation and training benchmark models on synthetic multilingual dialogs as well as multilingual transcripts.  }
    \label{fig:pipeline}
\end{figure*}

The overall dataset creation process consists of five core steps: 1) definition of the annotation schema: dialog-related personally identifiable information, 2) generation of synthetic dialogs in English, 3) curation and annotation of sensitive content in synthetic dialogs, 4) translation and curation of dialogs, and 5) verbalization, as well as transcription of dialogs with curation. An overview of the processing is presented in Figure \ref{fig:pipeline}. 

\subsection{Step 1: Definition of Annotation Schema }


We developed a dialog-oriented annotation schema that combines concepts from PHI-based de-identification frameworks with categories from the Text Anonymization Benchmark (TAB) \cite{pilan2022text}. Compared to traditional PHI schemas, TAB includes broader entity categories relevant to protect privacy beyond clinical documentation, such as organizations, locations and demographic attributes. Building upon these approaches, our schema was adapted specifically for conversational data and extended with additional categories frequently occurring in spoken interactions, including professions, social relationships, and product references. The resulting annotation schema contains categories covering persons, organizations, locations, identification codes, professions, products, temporal information, quantities, and miscellaneous identifying characteristics. 
Several categories were further divided into more fine-grained subcategories, such as phone numbers, email addresses, postal codes, or social relations. The complete annotation schema and definitions are presented in Table \ref{tab:annotation_schema}.

\begin{table*}[!htbp]
\centering
\footnotesize
\begin{tabular}{p{0.25\textwidth} p{0.68\textwidth}}
\toprule
\textbf{Entities+Subcategories} & \textbf{Definition} \\ 
\midrule
\begin{tabular}[t]{@{}l@{}}
PERSON \\ - EMAIL \\ - SOCIAL\_RELATION
\end{tabular}
& Names, user/nicknames, aliases, emails and initials. Merge academic titles and job titles with names (e.g., ``Dr.\ Anna Müller'', ``Sergeant Green''). Annotate also titles if they occur without the name (e.g.\ ``Frau Doktor'', ``Officer''). If the title is also a job description, annotate as PERSON only when someone is addressed specifically. SOCIAL\_RELATION includes all mentions of personal relations such as family members (e.g., ``father'', ``daughter'', ``cousin''), partners (e.g., ``husband'', ``girlfriend''), friends (e.g., ``best friend''), and professional relations (such as ``colleague'', ``manager'', ``boss''). \\ 
\midrule
\begin{tabular}[t]{@{}l@{}}
CODE \\ - PHONE \\ - URL
\end{tabular}
& Numbers and identification codes, such as social security numbers, phone numbers, passport numbers, license plates, account numbers, IP addresses, etc. Phone also includes fax. \\ 
\midrule
\begin{tabular}[t]{@{}l@{}}
LOCATION (LOC-OTHER) \\ - STREET \\ - HOUSE\_NUMBER \\ - CITY \\ - ZIP \\ - COUNTRY
\end{tabular}
& Places and locations, such as cities, areas, countries, addresses, named infrastructures, etc. \\ 
\midrule
ORGANIZATION (ORG) & Names of organizations, such as public and private companies, schools, universities, public institutions, prisons, healthcare institutions, non-governmental organizations, churches, etc. \\ 
\midrule
PROFESSION & All professions, employment status and sociodemographic information such as being a student, a prisoner, or being retired. This includes hobby professions, such as ``goalkeeper'', and explicit descriptions of professions such as ``driving a taxi''. Annotate e.g.\ ``[retail] job'', ``I work in [HR]''. \\
\midrule
PRODUCT & Complementary to ORG we introduce PRODUCT which includes all mentions of proper product names such as ``Dell XPS 15'' $\rightarrow$ ORG PRODUCT. Includes product types such as insurance policies, smartphone or laptop models, vehicles, medication, applications, etc. \\ 
\midrule
\begin{tabular}[t]{@{}l@{}}
DATETIME \\ - AGE
\end{tabular}
& Description of a specific date (e.g., October 3, 2018), time (e.g., 9:48 AM), duration (e.g., 18 years), or regularities (e.g.\ weekly, once per month). Includes mentions of weekdays (including ``weekend''), seasons (e.g.\ ``last [summer]''), and time of day (e.g., ``afternoon''). Do not annotate relational temporal information, such as ``a couple days ago'', or ``yesterday''. Mentions of temporal information that falls under QUANTITY should be annotated as DATETIME as well. Annotate e.g.\ ``[seven weeks] ago''. \\ 
\midrule
QUANTITY & Description of a meaningful quantity, e.g., percentages or monetary values, but also enumerations (``third operation''). Always annotate the measurement into the entity (e.g.\ ``three pints of beer''). Unspecific quantifiers such as ``a few'', ``all'', or ``some'' should not be annotated as QUANTITY. Mentions of QUANTITY that fit other categories (such as ``three siblings'' as SOCIAL\_RELATION, or ``seven weeks'' as DATETIME) should be annotated under the respective categories of higher priority. \\ 
\midrule
MISC & Any other characteristic that could uniquely identify the individual and that does not belong to the categories above. \\

\bottomrule

\end{tabular}
\caption{Overview of annotation schema with the different entities and subcategories, together with their definition.}
\label{tab:annotation_schema}
\end{table*}

The schema was designed to capture information that could plausibly contribute to re-identification with limited contextual knowledge. However, the schema does not aim to guarantee complete anonymization. In particular, indirect identifiers and highly context-dependent information were only partially considered, as their annotation is substantially more ambiguous and difficult to perform consistently across languages and conversational settings. Instead, the annotation guidelines were intentionally designed to balance practical applicability, annotation consistency, and coverage of highly relevant sensitive information. The development of the schema was conducted iteratively alongside the creation of the synthetic dialogs. During early dialog generation experiments, additional recurring forms of identifying information emerged that were insufficiently represented in existing de-identification frameworks. These observations influenced both the refinement of annotation categories and the selection of dialog scenarios included in the dataset. 

\subsection{Step 2: Synthetic Dialog Generation}

Dialogs were generated synthetically using the Gemini large language model (2.5 Pro) and subsequently manually curated to improve plausibility, linguistic naturalness, and diversity where necessary. Synthetic generation was selected as real-world conversational data containing sensitive information are difficult to access and typically subject to ethical, legal, and institutional restrictions. In addition, publicly shareable multilingual dialog resources for conversational de-identification remain scarce. To increase linguistic and contextual diversity, dialogs were generated across multiple interaction scenarios representing different situations in which sensitive information may naturally occur. Different conversational settings lead to different disclosure patterns and entity distributions. For example, emergency calls frequently contain location descriptions and phone numbers, whereas therapy sessions more often include social relationships, professions, or temporal references. The selected scenarios were therefore chosen to cover a broad spectrum of conversational contexts relevant to privacy-preserving natural language processing. 

Overall, the dataset contains seven primary dialog scenarios with 20 dialogs each. Most dialogs involve two speakers engaged in task-oriented or semi-structured interactions. In addition, seven extended multi-speaker group therapy conversations were generated. Compared to the remaining scenarios, these group discussions are substantially longer and contain more complex interaction dynamics, including overlapping personal narratives and references to multiple participants. 

%
%
%
%
%
%
%
%
%
%

\begin{table*}[htbp]
    \centering
    \includegraphics[width=0.9\textwidth]{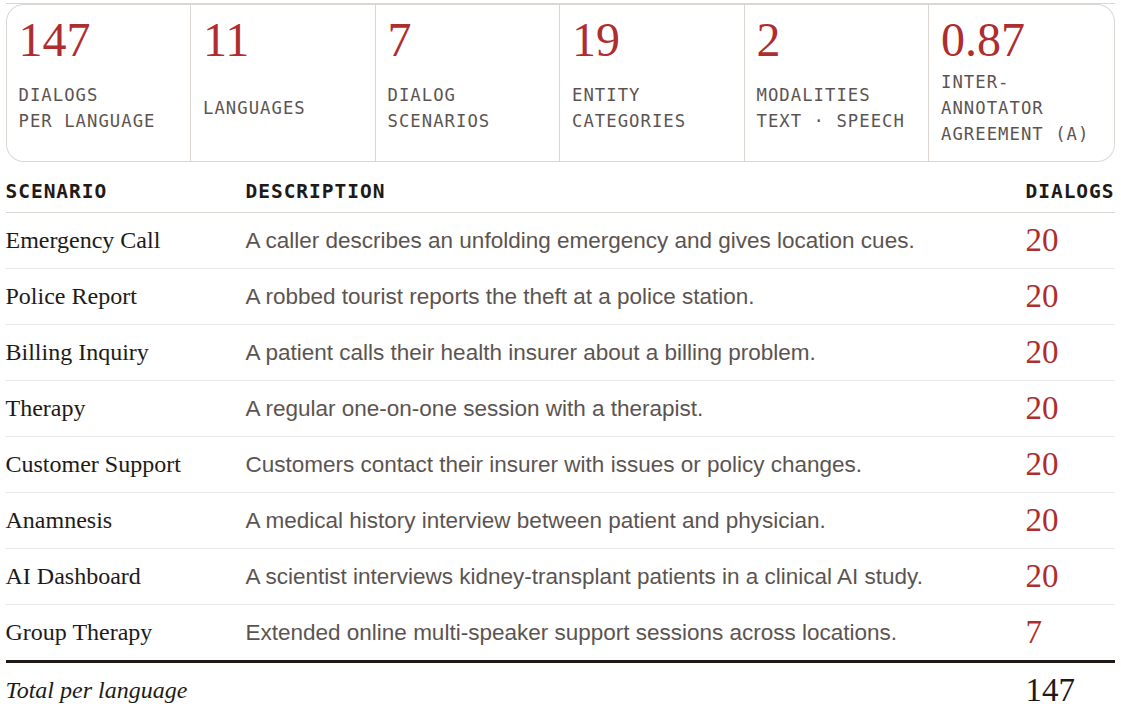}
    \caption{Overview of different dialog setups}
    \label{tab:use_cases}
\end{table*}

Dialogs were generated using structured prompts defining the conversational scenario, interaction goals, speaker personas, and geographic context. Prompts specified information such as speaker age, occupation, personality traits, and conversational setting to encourage variation in speaking style and entity usage. Each dialog scenario was associated with one or more predefined cities within the United Kingdom, primarily Sheffield, London, and Manchester. Group therapy sessions were not restricted to a single geographic location and included participants from multiple regions. The inclusion of multiple cities was motivated by preliminary experiments showing that large language models tend to repeatedly generate identical locations, names, institutions, and conversational patterns over time. Varying the geographic context therefore increased diversity in street names, organizations, healthcare institutions, and culturally specific references. While city selection was manually controlled, persona generation and scenario instantiation were performed semi-automatically using the language model. 

During early iterations of dialog generation, repetitive dialog structures and homogeneous entity distributions were frequently observed despite the use of different personas and prompts. To mitigate these effects, prompts were iteratively refined with additional instructions encouraging diversity in interaction style, demographic background, conversational flow, and disclosure behavior. Dialogs were subsequently manually reviewed and adjusted where necessary to reduce repetitive artifacts and implausible conversational patterns. An example prompt with exemplary scenario, is presented in the Appendix \ref{prompt_dialog_generation}.

\subsection{Step 3: Curation and Annotations}

Following dialog generation, all conversations were manually reviewed to improve linguistic quality, plausibility, and diversity. Dialogs containing highly unnatural, logically inconsistent, or repetitive conversational patterns were revised manually. In particular, repetitive artifacts generated by the language model occurred frequently during early iterations, including recurring names, locations, institutions, and highly similar interaction structures across dialogs. These issues were manually corrected to increase variability across scenarios and speakers. Each dialog was reviewed by at least one annotator prior to annotation. 

\begin{table}[!h]
\centering
\setlength{\tabcolsep}{3pt}
\begin{tabular}{lccccccccc}
\toprule
ENTITY & \rotatebox{90}{OVERALL} & \rotatebox{90}{AI-DASHBOARD} & \rotatebox{90}{ANAMNESIS} & \rotatebox{90}{BILLING} & \rotatebox{90}{CAR-INSURANCE} & \rotatebox{90}{GROUP-THERAPY} & \rotatebox{90}{POLICE} & \rotatebox{90}{THERAPY} & \rotatebox{90}{TRIAGE} \\
\midrule
PERSON & 3044 & 408 & 454 & 690 & 432 & 276 & 202 & 486 & 96 \\
- EMAIL & 246 & 0 & 0 & 36 & 194 & 0 & 16 & 0 & 0 \\
- SOCIAL\_RELATION & 1152 & 122 & 314 & 80 & 6 & 308 & 110 & 178 & 34 \\
CODE & 444 & 4 & 44 & 112 & 246 & 0 & 38 & 0 & 0 \\
- PHONE & 90 & 2 & 36 & 6 & 6 & 0 & 20 & 0 & 20 \\
- URL & 10 & 0 & 0 & 0 & 10 & 0 & 0 & 0 & 0 \\
LOCATION (OTHER) & 86 & 4 & 0 & 4 & 0 & 0 & 22 & 44 & 12 \\
- STREET & 178 & 8 & 46 & 10 & 6 & 0 & 34 & 24 & 50 \\
- HOUSENUMBER & 80 & 0 & 38 & 2 & 4 & 0 & 10 & 0 & 26 \\
- CITY & 514 & 42 & 128 & 32 & 10 & 76 & 92 & 100 & 34 \\
- ZIP & 54 & 0 & 26 & 4 & 4 & 0 & 6 & 0 & 14 \\
- COUNTRY & 172 & 16 & 2 & 4 & 0 & 8 & 132 & 8 & 2 \\
ORG & 1156 & 46 & 134 & 538 & 124 & 4 & 154 & 146 & 10 \\
PROFESSION & 934 & 192 & 182 & 326 & 26 & 82 & 28 & 94 & 4 \\
PRODUCT & 312 & 8 & 116 & 72 & 60 & 2 & 44 & 0 & 10 \\
DATETIME & 1074 & 92 & 368 & 230 & 94 & 38 & 82 & 158 & 12 \\
- AGE & 176 & 20 & 88 & 4 & 0 & 16 & 20 & 12 & 16 \\
QUANTITY & 334 & 28 & 126 & 58 & 32 & 20 & 28 & 32 & 10 \\
MISC & 28 & 8 & 6 & 2 & 0 & 2 & 2 & 8 & 0 \\
\midrule
\textbf{TOTAL} & \textbf{10084} & \textbf{1000} & \textbf{2108} & \textbf{2210} & \textbf{1254} & \textbf{832} & \textbf{1040} & \textbf{1290} & \textbf{350} \\
\bottomrule
\end{tabular}
\caption{Overview of annotated entities across dialog scenarios}
\label{tab:entites_scenarios}
\end{table}

Sensitive information annotation was conducted using the browser-based annotation platform INCEpTION \cite{klie-etal-2018-inception}. Dialogs were annotated according to the annotation schema described in Step 1. Each dialog was independently annotated by two annotators. In cases of disagreement, annotations were reviewed jointly by one of the original annotators and an additional annotator to produce the final gold-standard annotations. Annotations were performed at entity-span level and included all categories defined in the annotation schema, including person-related entities, organizations, locations, identification codes, temporal expressions, professions, products, quantities, and miscellaneous identifying characteristics. Table \ref{tab:entites_scenarios} summarizes the distribution of annotated entities across different dialog scenarios. While some information is generally rare, such as STATE, the table shows that use cases might elicit different entity types. The annotation guidelines are available on Zenodo\footnote{\label{fn:zenodo}\url{https://zenodo.org/records/20863452}}.  
\newcounter{zenodofootnote}
\setcounter{zenodofootnote}{\value{footnote}}

\subsection{Step 4: Translations and Curation}

The curated English dialogs were subsequently translated into ten additional languages: Arabic, Finnish, French, German, Hindi, Italian, Polish, Portuguese, Spanish and Turkish. Translations were generated using the Gemini large language model (2.5 Pro) and localized to country-specific and city-specific settings to improve cultural and linguistic realism. Rather than performing direct literal translations, dialogs were adapted to the corresponding sociocultural context of each language. For example, dialogs originally located in London were transformed into culturally plausible settings such as Berlin for German dialogs or Milan for Italian dialogs. Localization included adaptation of names, streets, institutions, healthcare systems, and culturally specific conversational references where appropriate. Cities were selected manually based on familiarity and linguistic confidence of the annotators to facilitate culturally plausible revisions and quality control. The prompt to translate the dialogs is presented in Appendix \ref{prompt_translate}. Table \ref{tab:lang_cities} provides an overview of the languages, countries, and geographic settings used throughout the dataset.  The guidelines for the annotators are available on Zenodo\textsuperscript{\hyperref[fn:zenodo]{\ref*{fn:zenodo}}}.

\begin{table*}[htbp]
    \centering
    \includegraphics[width=0.7\textwidth]{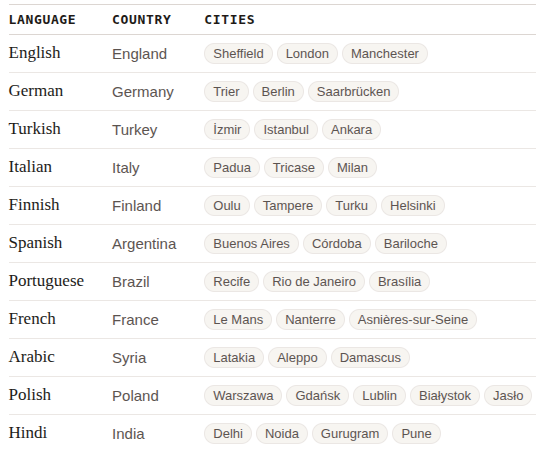}
    \caption{Overview of languages, the countries, and the corresponding cities in which dialogs take place. Texts for each language are situated in different countries, including 3-5 different cities. The cities were places the annotators felt confident with, as they knew the places.}
    \label{tab:lang_cities}
\end{table*}
%
%
%
%
%
%
%
%
%
%
%
%
%

After translation, all dialogs underwent an additional manual curation step. Annotators reviewed each dialog to identify and correct: 1) linguistic and grammatical errors, 2) culturally implausible or contextually inconsistent content, 3) incorrect or incomplete transferred annotations, and 4) repetitive or homogeneous entity usage introduced during translation. Special attention was given to localization quality, including culturally appropriate names, institutions, address formats, and conversational conventions. In several languages, additional modifications were introduced to reduce repetitive patterns caused by the language model, such as recurring names, locations, or dialog structures. 

The automatic annotation transfer from English to the target languages was additionally reviewed and corrected manually where necessary. A final consistency pass was performed semi-automatically by one annotator to harmonize annotation behavior across languages. Due to linguistic differences and localized modifications of dialog content, annotation distributions differ slightly between languages. Table \ref{tab:entities_across_lang} summarizes the final annotation statistics across all languages.

\begin{table}[h]
\centering
\footnotesize
\setlength{\tabcolsep}{3pt}
\begin{tabular}{lcccccccccccc}
\toprule
ENTITY & EN & AR & DE & FI & FR & HI & IT & PL & PT & SP & TR \\
\midrule
PERSON & 3060 & 3024 & 3054 & 3003 & 3048 & 3006 & 3046 & 3047 & 3102 & 3054 & 3068 \\
- EMAIL & 246 & 214 & 246 & 237 & 231 & 258 & 241 & 242 & 248 & 244 & 236 \\
- SOCIAL\_RELATION & 1154 & 1082 & 1150 & 1147 & 1158 & 1153 & 1150 & 1161 & 1172 & 1146 & 1180 \\
\midrule
CODE & 444 & 472 & 446 & 435 & 442 & 465 & 437 & 452 & 454 & 446 & 459 \\
- PHONE & 92 & 70 & 94 & 93 & 92 & 85 & 93 & 84 & 90 & 88 & 109 \\
- URL & 10 & 10 & 10 & 10 & 10 & 10 & 10 & 10 & 10 & 10 & 10 \\
\midrule
LOCATION (OTHER) & 86 & 76 & 100 & 90 & 108 & 84 & 82 & 110 & 86 & 86 & 97 \\
- STREET & 184 & 125 & 170 & 171 & 170 & 180 & 179 & 184 & 175 & 174 & 178 \\
- CITY & 518 & 474 & 502 & 485 & 494 & 506 & 493 & 566 & 507 & 504 & 515 \\
- ZIP & 54 & 22 & 54 & 53 & 54 & 52 & 53 & 52 & 55 & 52 & 52 \\
- COUNTRY & 172 & 160 & 174 & 172 & 166 & 168 & 170 & 158 & 170 & 172 & 173 \\
\midrule
DATETIME & 1076 & 889 & 1000 & 912 & 886 & 923 & 966 & 918 & 916 & 978 & 832 \\
- AGE & 178 & 156 & 176 & 171 & 165 & 173 & 174 & 174 & 158 & 174 & 170 \\
\midrule
ORG & 1156 & 1084 & 1086 & 1068 & 1156 & 1066 & 1107 & 1122 & 1136 & 1100 & 1115 \\
PROFESSION & 934 & 858 & 934 & 935 & 862 & 931 & 876 & 1013 & 914 & 910 & 926 \\
PRODUCT & 312 & 308 & 320 & 318 & 324 & 318 & 315 & 310 & 320 & 310 & 338 \\
QUANTITY & 334 & 316 & 332 & 331 & 423 & 333 & 326 & 340 & 337 & 334 & 349 \\
MISC & 28 & 28 & 30 & 28 & 44 & 30 & 28 & 32 & 28 & 30 & 40 \\
\midrule
\textbf{TOTAL} & \textbf{10038} & \textbf{9368} & \textbf{9878} & \textbf{9659} & \textbf{9833} & \textbf{9741} & \textbf{9746} & \textbf{9975} & \textbf{9878} & \textbf{9812} & \textbf{9847} \\
\bottomrule
\end{tabular}
\caption{Overview of annotated entities across languages}
\label{tab:entities_across_lang}
\end{table}

\subsection{Step 5: Transcription, Post Processing and Curation}

To support research on speech anonymization and conversational automatic speech recognition, all written dialogs were converted into synthetic speech using the Google Cloud Text-to-Speech (TTS) API\footnote{\url{https://cloud.google.com/text-to-speech}}, utilizing Chirp (HD), Studio, and WaveNet voices.  Wherever possible, different synthetic voices were assigned to different speakers to preserve speaker distinctions within the conversations. Due to limited voice availability, this was not consistently possible for all languages, particularly Hindi. The generated audio files were subsequently transcribed using a locally deployed instance of Whisper \cite{bain2023whisperx} (WhisperX\_large\_v3) and Pyannote speaker diarization \cite{bredin2023pyannote} to differentiate between speaker turns. Resulting transcripts therefore contain realistic automatic speech recognition artifacts, including transcription errors, punctuation inconsistencies, and occasional speaker segmentation failures, which might make the detection of sensitive information more difficult. 

Following transcription, sensitive information annotations from the written dialogs were automatically projected onto the generated transcripts.
Projection was challenging because ASR errors could alter token sequences, entity boundaries, and punctuation, while speaker diarization occasionally failed to preserve the original turn structure.
Low-confidence projections and problematic spans were therefore reviewed and corrected manually.

\section{Data Record}

The dataset \textbf{DialogPII} is publicly available on Zenodo\footnote{\url{https://zenodo.org/records/20863452}} and organized by resource type, language, and dialog scenario. The repository contains synthetic dialogs, speech transcripts, sensitive information annotations, and associated metadata for 11 languages. 

\subsection{Repository Structure}

The dataset is divided into two primary resources: 
\begin{itemize}
    \item text-based synthetic dialogs with sensitive information annotations, 
    \item speech-derived transcripts generated through text-to-speech and automatic speech recognition pipelines. 
\end{itemize}

The repository is organized hierarchically by resource type (original dialog or transcript), language, and dialog scenario. 

\subsection{Synthetic Dialogs}

The \texttt{dialogs/} directory contains manually curated synthetic dialogs and corresponding annotations for sensitive information spans. 

Structure of the data directory: 

\begin{tcolorbox}[colback=white, colframe=black, boxrule=0.4pt, breakable]
\begin{Verbatim}
DialogPII/
|-- dialogs/
|   |-- AR/
|   |   |-- AR_aidashboard.json
|   |   |-- AR_anamnesis.json
|   |   |-- ...
|   |-- DE/
|   |-- EN/
|   |-- ...
|-- transcripts/
    |-- AR/     
    |   |-- AR_aidashboard_transcript.json
    |   |-- ...
    |-- ...
\end{Verbatim}
\end{tcolorbox}

Each language folder contains dialogs for seven predefined two-speaker scenarios, with 20 dialogs per scenario.
In addition, each language includes seven extended multi-speaker group-therapy dialogs.
These group dialogs are substantially longer and simulate multi-party therapeutic or self-help group interactions.
Each language-scenario file is stored in JSON format and contains all dialogs for that language and scenario, including:
\begin{itemize}
\item metadata describing language, scenario, and version,
\item dialog records with chat numbers,
\item turn-level speaker information,
\item turn text,
\item turn-local sensitive information annotations with category labels.
\end{itemize}

Annotation offsets are zero-based, start-inclusive, and end-exclusive. They are relative to the corresponding turn text.
An example record is shown below.
The example illustrates the transcript format; original dialogs use the same turn-based structure but omit timestamps.

\begin{tcolorbox}[colback=white, colframe=black, boxrule=0.4pt, breakable]
\begin{verbatim}
{
  "language": "EN",
  "scenario": "therapy",
  "version": "transcript",
  "dialogs": [
    {
      "chat_number": 4,
      "dialog": {
        "turns": [
          {
            "speaker": "SPEAKER_00",
            "start_time": 12.34,
            "end_time": 18.76,
            "text": "When Sophie moved to Manchester, ...",
            "annotations": [
              {
                "type": "PERSON",
                "start": 5,
                "end": 11,
                "text": "Sophie"
              } ...
\end{verbatim}
\end{tcolorbox}

\subsection{Speech-Based Transcripts}

The \texttt{transcripts/} directory contains speech-derived versions of the dialogs. These resources were generated to support research on speech anonymization and sensitive information detection in automatic speech recognition pipelines. 

For each dialog: 

\begin{itemize}
    \item text dialogs were converted to speech using Text-to-Speech (TTS),
    \item generated audio was transcribed using Whisper, 
    \item sensitive information annotations were automatically projected onto transcripts, 
    \item projected annotations were manually curated. 
\end{itemize}

The directory structure mirrors the organization of the original dialogs.
As with the original dialogs, each transcript record contains:

\begin{itemize}
\item turn-level ASR transcript text,
\item speaker segmentation,
\item projected and curated annotations,
\item metadata identifying the corresponding original dialog.
\end{itemize}

\subsection{Languages}

The dataset contains dialogs and transcripts in the following 11 languages: 

\begin{itemize}
\item Arabic (AR)
\item German (DE)
\item English (EN)
\item Finnish (FI)
\item French (FR)
\item Hindi (HI)
\item Italian (IT)
\item Polish (PL)
\item Portuguese (PT)
\item Spanish (SP)
\item Turkish (TR)
\end{itemize}

Dialogs were localized to country- and city-specific settings to increase linguistic and cultural variation across scenarios. 

\subsection{Data Statistics}

Overall, the dataset contains 147 dialogs per language, comprising: 

\begin{itemize}
    \item 7 dialog scenarios, 
    \item 20 dialogs per scenario, 
    \item 7 additional multi-speaker group dialogs, 
    \item annotated sensitive information spans, 
    \item speech-derived transcripts for all dialogs across all languages. 
\end{itemize}

Detailed statistics regarding annotation categories, dialog lengths, and transcript properties are provided in Tables 1-5.

\section{Technical Validation}

\subsection{Annotation Quality }

The annotation of the English dialogs was conducted by two annotators. Disagreements were resolved by a team of one of the original annotators and a third annotator who was trained in using the schema, resulting in the final annotations. The inter-annotator agreement between the two annotators was 0.87 Krippendorff’s alpha unitizing. 

\subsection{Translation Quality }

\paragraph{Automatic Quality Estimation:} Each translated dialog was reviewed by at least one person, correcting/modifying the setup and dialog, for instance if the text was not linguistically correct or not natural, to improve the dialog to the new context of the country, or to provide more variations. Moreover, automatically generated (translated) annotations had to be revised and corrected if necessary. Table \ref{tab:edits_lang} shows an overview of edits per language as well as modified annotations. 

\begin{table}[h]
\centering
\setlength{\tabcolsep}{3pt}
\begin{tabular}{lcccccccccc}
\toprule
 & AR & DE & FI & FR & HI & IT & PL & PT & SP & TR  \\ 
\midrule
Avg Edits (\%)  & 5.4 & 7.9  & 4.1 & 0.8 & 9.2  & 1.4  & 1.7  & 0.6  & 1.2  & 2.6  \\ 
Insert  & 10  & 275  & 104 & 6 & 19  & 26  & 16  & 16  & 132  & 13  \\ 
Deletions  & 23  & 172  & 122  & 7  & 7  & 21  & 13  & 20  & 33  & 43  \\ 
Replacements  & 4158  & 7763  & 3187  & 812  & 10464  & 1268  & 1486  & 587  & 1014  & 1144 \\ 
\bottomrule
\end{tabular}
\caption{Overview of edits and changes per language}
\label{tab:edits_lang}
\end{table}

\subsection{Manual Quality Estimation}

As described in the following, annotators were rating the translations according to different aspects on a Likert scale from 1-5 (1 = very poor / strongly disagree, 5 = excellent / strongly agree). 

\begin{table*}[htbp]
    \centering
    \includegraphics[width=0.7\textwidth]{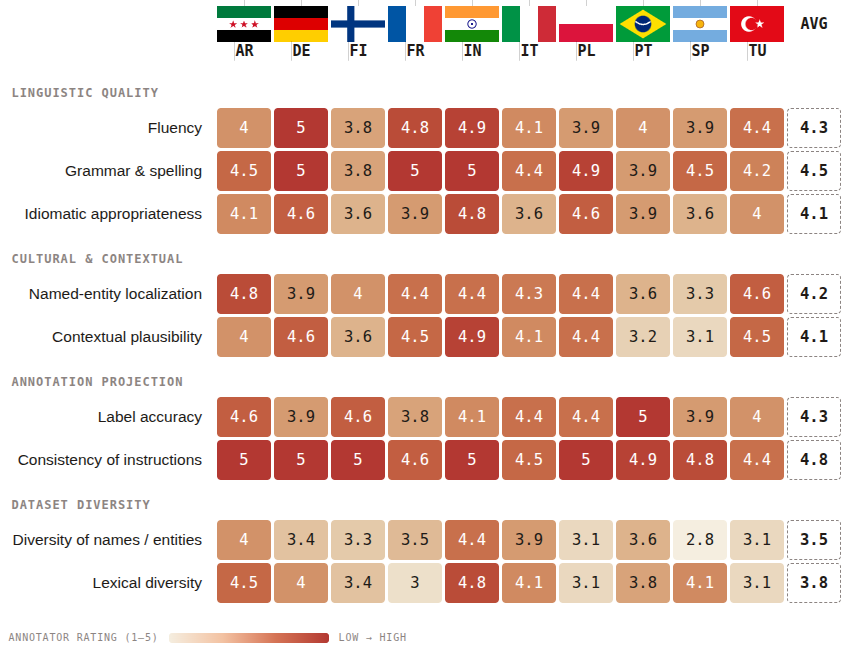}
    \caption{Overview of data quality of translations regarding different dimensions: linguistic quality, cultural and contextual, annotation projection and dataset diversity. }
    \label{tab:data_quality}
\end{table*}

\paragraph{Translation Quality:} Table \ref{tab:data_quality} depicts that the translations were perceived as linguistically strong, with average scores of 4.3 for fluency and 4.5 for grammar and spelling. Annotators frequently noted that the texts were generally understandable and grammatically correct, but many dialogs sounded overly literal or too close to the English source structure. This was particularly mentioned for Turkish, Finnish, Polish, and Arabic, where direct translations resulted in unnatural phrasing, unusual sentence order, or expressions that native speakers would rarely use. Several languages also showed excessive formality or inconsistent register usage, such as switching between formal and informal pronouns in German, Spanish, and French dialogs. 

Another recurring issue concerned cultural and contextual localization. While named entity localization received relatively high scores overall (4.2), annotators often pointed out implausible or repetitive scenarios, unrealistic tourist behavior (police use case), and culturally inappropriate details. Examples included foreign tourists speaking flawless local language varieties, unrealistic healthcare or insurance references, repeated use of the same names and locations, and culturally uncommon references such as alcohol consumption patterns in Arabic dialogs. Address formatting, time expressions, greetings, and institution references also frequently reflected English conventions rather than local norms. 

Despite these issues, reviewers generally appreciated that many entities, locations, and contextual references had been adapted to the target languages. Some annotators specifically noted that translations felt realistic after correction, particularly in Hindi, French, and Arabic, although several datasets still suffered from repetitive dialog structures and limited situational diversity.

%

\paragraph{Annotation Projection Quality:} As presented in the lower part of Table \ref{tab:data_quality}, the annotation projection quality was rated positively overall, with an average label accuracy score of 4.3 and very high consistency scores for the annotation guidelines (4.8). Nevertheless, annotators across nearly all languages reported systematic inconsistencies in entity projection. Common problems included missing tags, incorrect label assignments, broken XML tags, and entities being annotated differently across dialogs. Several reviewers also observed that the projection often followed English annotation boundaries too closely, leading to unnatural spans in morphologically rich languages such as Finnish, Polish, and Arabic. 

Many annotation errors involved confusion between semantically related categories, such as organizations (e.g. embassy) being tagged as locations. Annotators also highlighted issues with partial entity annotation, for example when suffixes in agglutinative languages were excluded from entity spans, or when titles such as ``Herr'', or ``Frau'' were incorrectly merged into person entities. In a few cases, annotators encountered entirely new or invalid labels such as PLACE, ACTIVITY, or IDIOM, indicating occasional instability in the annotation pipeline. 

At the same time, reviewers generally agreed that the annotation guidelines themselves were clear and internally consistent. Most corrections therefore focused on projection quality rather than ambiguity in the task instructions. The relatively high manual ratings suggest that, although projection artifacts were frequent, annotators were usually able to identify and correct them efficiently.

%
%

\paragraph{Dataset Diversity:} As shown in the lower part of Table \ref{tab:data_quality}, the dataset diversity received the weakest ratings overall, especially regarding diversity of names and entities (average 3.5). Across languages, annotators repeatedly commented on repetitive dialog structures, recurring scenarios, and overuse of the same names, organizations, hospitals, and locations. Several datasets reused the same districts, surnames, or institutions across multiple dialogs, making the generated conversations feel templated. This issue was especially noted in German, Finnish, Spanish, Portuguese, and Turkish. 

Annotators also observed limited demographic and cultural diversity. Dialogs frequently depicted heteronormative families, stereotypical tourist situations, or implausible combinations of nationalities and language abilities. Some comments mentioned that certain minority groups or cultural references appeared disproportionately often relative to local realities, while other common groups or situations were largely absent. Repetition of phone brands, passport formats, insurance providers, and medical conditions further contributed to a sense of artificiality. 

Lexical diversity was evaluated somewhat more positively (average 3.8), although many reviewers still noted repetitive filler expressions, overused slang, and recurring English discourse patterns. Some languages, such as French and Turkish, were criticized for having nearly identical dialog structures across scenarios, while others, including Hindi and Arabic, were viewed as somewhat more varied after localization corrections. Overall, the feedback suggests that future dataset generation would benefit from stronger diversification strategies for names, settings, interaction styles, and sociocultural contexts. 

%
%
%
%

\paragraph{Quality of annotations in transcriptions }

Entity annotations from the original dialogs were automatically projected onto the ASR-derived transcripts. Annotations that could not be mapped with high confidence were subsequently reviewed and corrected manually. Across all eleven languages, 3,834 of 53,057 annotations (7.23\%) were initially flagged as review items. The number of automatically mapped items was relatively stable across languages. However, the proportion of items requiring review varied substantially. Most languages had review rates between 4 and 7\%, while Hindi and Arabic showed considerably higher rates, with 14.99\% and 14.02\%, respectively. 

An additional quality control pass was conducted to correct malformed or inconsistent tags introduced by the automatic translation of the original dialogs. These issues were identified during the review process of the automated annotation mapping. We detected 2,564 tag-level issues in 265 source files that were resolved through a combination of targeted automatic repairs and manual adjudication in both the original files and their respective transcripts.

\subsection{Baseline Models }

The main focus of this work is the presentation of DialogPII, a shareable, multilingual resource to detect sensitive information in dialog transcripts.  Along with this data, we demonstrate the utility of the dataset for multilingual sensitive information detection. For this reason, we present a baseline named entity recognition model on the curated dialog corpus, first on a) synthetic dialogs and then on b) speech-transcribed data. We fine-tuned a multilingual transformer-based model using jointly combined training data from all 11 languages. The model is based on the ModernBERT \cite{warner2025smarter} architecture and extended with a Conditional Random Field (CRF) decoding layer for sequence labeling. 

\paragraph{Model Architecture }

\begin{itemize}
    \item \textbf{Base model}: mmBERT-base \cite{marone2025mmbertmodernmultilingualencoder}
    \item \textbf{Sequence labeling head}: linear classification layer with CRF decoding 
    \item \textbf{Decoding}: Viterbi decoding 
    \item \textbf{Training objective}: token-level BIO tagging 
    \item \textbf{Context enrichment}: document-level features via FLERT \cite{schweter2020flert} - the surrounding 2 sentences are encoded alongside the target sentence to provide cross-sentence context
\end{itemize}

The model predicts 19 categories of sensitive information, including person names, addresses, organizations, phone numbers, URLs, and temporal expressions. 

\paragraph{Training Configuration} 

The model was trained jointly on all languages. Each batch was constructed to include examples from all 11 languages. Table \ref{tab:hyperparameters} depicts the hyperparameters used in our experiments. 

\begin{table}[h]
\centering
\begin{tabular}{lc}
\toprule
\textbf{Parameter} & \textbf{Value} \\ 
\midrule
Learning rate & 2e-5 \\ 
Batch size  & 32 \\ 
Maximum sequence length & 2048 \\ 
Dropout & 0.1 \\ 
Epochs & 10 \\ 
\bottomrule
\end{tabular}
\caption{Hyperparameters for training}
\label{tab:hyperparameters}
\end{table}

\paragraph{Evaluation}

To demonstrate the suitability of DialogPII for multilingual sensitive information detection, we trained a baseline multilingual named entity recognition model and evaluated it on held-out test sets for each language. Performance was measured using micro-averaged F1 scores under two evaluation settings. First, we report exact span matching, which requires predicted entities to exactly match the gold-standard start and end boundaries. Second, we report lenient span matching, where partially overlapping entity spans are counted as correct. Since de-identification systems are often primarily concerned with detecting potentially identifying information regardless of fine-grained category assignment, we additionally report type-agnostic (TA) scores. Under this setting, only the detection of a sensitive span is evaluated, while the predicted entity category is ignored. For example, a ZIP code incorrectly classified as a generic code entity would still be considered correct if the sensitive span itself was detected.

\begin{table}[h]
  \small
  \setlength{\tabcolsep}{3pt}
  \begin{tabular}{lcccccccccccc}
    \toprule
    & {AR} & {DE} & {EN} & {FI} & {FR} & {HI} & {IT} & {PL} & {PT} & {SP} & {TR} & {Avg} \\
    \midrule
    {Lenient F1} & 79.84 & 92.36 & 94.18 & 89.89 & 89.48 & 84.86 & 90.60 & 90.21 & 90.90 & 92.26 & 87.81 & 89.31 \\
    {Exact F1}  & 76.73 & 91.58 & 91.69 & 88.48 & 86.28 & 80.81 & 87.84 & 87.51 & 89.08 & 90.51 & 84.52 & 86.82 \\
    \midrule
    {L. F1 (TA)} & 82.89 & 94.11 & 95.14 & 92.55 & 91.30 & 88.10 & 91.57 & 92.27 & 92.15 & 93.75 & 89.97 & 91.25 \\
    {E. F1 (TA)}  & 79.64 & 92.98 & 92.58 & 90.99 & 87.74 & 83.59 & 88.68 & 89.22 & 90.10 & 91.89 & 86.32 & 88.52 \\
    \bottomrule
  \end{tabular}
    \caption{Base model trained and evaluated on synthetic dialogs. The table presents lenient (L.) and exact (E.) micro F1 scores across languages. The lower part of the table presents Type-Agnostic (TA) scores, which ignore the predicted entity type, it evaluates only whether a sensitive span was detected.}
  \label{tab:len-exf1-transposed}
\end{table}

\begin{table}[h]
  \centering
  \small
  \setlength{\tabcolsep}{3pt}
  \begin{tabular}{l *{11}{S[table-format=2.2]} S[table-format=2.2]}
    \toprule
    & {AR} & {DE} & {EN} & {FI} & {FR} & {HI} & {IT} & {PL} & {PT} & {SP} & {TR} & {Avg} \\
    \midrule
    {Lenient F1} & 69.10 & 91.41 & 91.94 & 85.79 & 88.21 & 80.67 & 89.29 & 88.50 & 86.69 & 87.22 & 82.10 & 85.54 \\
    {Exact F1}  & 61.94 & 89.96 & 88.59 & 83.19 & 83.98 & 75.34 & 86.20 & 84.79 & 83.24 & 84.58 & 78.07 & 81.81 \\
    \midrule
    {L. F1 (TA)} & 71.52 & 93.27 & 93.58 & 88.08 & 89.93 & 84.61 & 90.63 & 90.57 & 88.08 & 88.88 & 85.24 & 87.67 \\
    {E. F1 (TA)}  & 63.59 & 91.49 & 89.98 & 84.90 & 85.15 & 78.04 & 87.03 & 86.18 & 84.17 & 85.86 & 80.74 & 83.37 \\
    \bottomrule
  \end{tabular}
  \caption{Base model trained and evaluated on speech-transcribed dialogs. The table presents lenient (L.) and exact (E.) micro F1 scores across languages. The lower part of the table presents Type-Agnostic (TA) scores.}
  \label{tab:test_transcript}
\end{table}

Table \ref{tab:len-exf1-transposed} presents the results obtained when training and evaluating the model on the curated synthetic dialogs. Overall performance was high across most languages, with an average lenient F1 score of 89.31 and an average exact F1 score of 86.82. English, German, Spanish, Portuguese, and Italian achieved lenient F1 scores above 90, indicating that the dataset supports effective multilingual learning despite substantial variation in language families and dialog scenarios. Performance was lower for Arabic and Hindi, reflecting the additional challenges posed by different scripts, morphological variation, and lower-resource language settings.

Table \ref{tab:test_transcript} reports results obtained on the speech-transcribed version of the dataset. Compared to the original dialog data, performance decreased moderately across most languages. The average lenient F1 score decreased from 89.31 to 85.54, while the average exact F1 score decreased from 86.82 to 81.81. This reduction is expected, as automatic speech recognition introduces transcription errors that may alter entity boundaries or affect entity recognition. Nevertheless, the model maintained strong performance across the majority of languages, demonstrating that the speech-derived resources constitute a viable benchmark for studying de-identification in automatic speech recognition pipelines.

\paragraph{External Validation on Real-World Conversations}

As DialogPII consists of curated, synthetic dialogs, an important question is whether models trained on the dataset generalize beyond the generated scenarios. To investigate this, we performed an external evaluation using a subset of the CallFriend corpus \cite{canavan1996callfriend,yaegerdror2006callfriend}. We selected five English and five German telephone conversations and manually annotated their transcripts according to the DialogPII annotation guidelines. The CallFriend conversations differ substantially from the scenarios represented in DialogPII. They consist of spontaneous telephone conversations between friends and family members discussing everyday topics and therefore contain more informal language, conversational digressions, and naturally occurring speech phenomena.

Tables \ref{tab:callfriend_basemodel_syntheitcdialogs} and \ref{tab:callfriend_basemodel_transcripts} present the results obtained using the baseline models trained on the synthetic dialogs and speech-transcribed dialogs, respectively. As expected, performance was lower than on the in-domain DialogPII test sets. Nevertheless, both models achieved encouraging results, particularly under the type-agnostic evaluation setting, where macro-average lenient F1 scores exceeded 83.

Several factors may contribute to the observed performance gap. First, the conversational domains represented in CallFriend differ considerably from the targeted scenarios included in DialogPII, resulting in different frequencies and distributions of sensitive entities. Second, real-world conversations contain greater linguistic variability, disfluencies, and spontaneous speech phenomena than curated synthetic dialogs. Third, certain entity formats, such as telephone numbers or postal codes, differ from the patterns represented in the synthetic data. Despite these limitations, the results suggest that models trained on DialogPII learn transferable representations of conversational sensitive information and can generalize beyond the synthetic training environment. The benchmark should therefore be viewed as evidence that the dataset provides a useful foundation for developing multilingual de-identification systems, while leaving substantial room for future work on domain adaptation and robustness to naturally occurring conversational data.

\begin{table}[h]
  \centering
  \small
  \begin{tabular}{l S[table-format=2.2] S[table-format=2.2] S[table-format=2.2] S[table-format=2.2]}
    \toprule
    & \multicolumn{2}{c}{\textbf{Normal}} & \multicolumn{2}{c}{\textbf{Type Agnostic}} \\
    \cmidrule(lr){2-3} \cmidrule(lr){4-5}
    \textbf{Language} & \textbf{Exact F1} & \textbf{Lenient F1} & \textbf{Exact F1} & \textbf{Lenient F1} \\
    \midrule
    English  & 69.70 & 71.86 & 82.13 & 84.95 \\
    German   & 73.66 & 76.02 & 82.80 & 85.79 \\
    \midrule
    \textbf{Macro avg} & 71.68 & 73.94 & 82.47 & 85.37 \\
    \bottomrule
  \end{tabular}
  \caption{Base model(2) evaluated on CallFriend dataset}
  \label{tab:callfriend_basemodel_syntheitcdialogs}
\end{table}

\begin{table}[h]
  \centering
  \small
  \begin{tabular}{l S[table-format=2.2] S[table-format=2.2] S[table-format=2.2] S[table-format=2.2]}
    \toprule
    & \multicolumn{2}{c}{\textbf{Normal}} & \multicolumn{2}{c}{\textbf{Type Agnostic}} \\
    \cmidrule(lr){2-3} \cmidrule(lr){4-5}
    \textbf{Language} & \textbf{Exact F1} & \textbf{Lenient F1} & \textbf{Exact F1} & \textbf{Lenient F1} \\
    \midrule
    English & 66.85 & 69.21 & 77.75 & 81.16 \\
    German  & 73.47 & 75.99 & 82.93 & 85.88 \\
    \midrule
    \textbf{Macro avg} & 70.16 & 72.60 & 80.34 & 83.52 \\
    \bottomrule
  \end{tabular}
  \caption{Transcript model(4) evaluated on CallFriend dataset }
  \label{tab:callfriend_basemodel_transcripts}
\end{table}

\section{Usage Notes}

DialogPII is intended for research on multilingual de-identification, conversational named entity recognition, speech anonymization, and privacy-preserving natural language processing. 

The repository provides aligned resources across text and speech modalities, allowing users to investigate: 

\begin{itemize}
    \item multilingual sensitive information detection,
    \item automatic speech recognition robustness, 
    \item annotation projection across modalities, 
    \item anonymization pipelines for conversational data.
\end{itemize} 

The released baseline models are optimized for conversational input and sentence-level inference. During model development, dialogs were segmented into sentence-level units prior to tokenization and prediction. Applying inference directly to long unsegmented speaker turns may reduce performance. 

Although the dialogs were manually curated, the dataset is synthetic and generated using large language models. Consequently, linguistic patterns and conversational dynamics may differ from naturally occurring human interactions. Researchers should therefore exercise caution when transferring models trained on this dataset to real-world clinical or customer-service environments. Performance differences across languages were observed, particularly for morphologically rich and lower-resource languages such as Arabic and Hindi.

\section{Data Availability }

The synthetic, as well as the transcribed dialogs, together with the annotation guidelines, are available on Zenodo\footnote{\url{https://zenodo.org/records/20863452}}.  

\section{Code Availability}

The code of the baseline models together with some exemplary code snippets is publicly available at Hugging Face\footnote{\url{https://huggingface.co/DFKI-SLT/multilingual_DialogPII_NER}}. Pretrained baseline checkpoints are distributed through Hugging Face and include configuration files required for CRF-based decoding.

\section{Acknowledgements }

This research was supported by the Federal Ministry of Research, Technology and Space (BMFTR) through the project Veranda (16KIS2046). 

\section{Author Contributions }

All: Read, reviewed and revised manuscript; RR: Idea, Planning, Conduction, Analysis; VCz: Planning, Conduction, Annotation, Curation, Coordination, Analysis; DE: Development of base models; IB, MP, MS, AJ, EG, VCo, MN, FB: Annotation, Curation; CH: Curation; AE: Support dataset; LF: Idea, Planning 

\section{Competing Interests }

No competing interests. 

\backmatter



\begin{appendices}

\section{Prompt to generate original dialogs}\label{prompt_dialog_generation}

To generate original English dialogs, an LLM was used in a first step to generate different example setups including personas (see example in Table \ref{tab:scenarios}), which were then inserted into a prompt to generate the complete dialogs, as presented in Figure \ref{fig:example_prompt_police}, for police. 

\begin{table}[h]
\centering
\begin{tabular}{p{0.68\textwidth}}
\toprule
A couple from Poland (32 and 30) were approached by a woman asking them to sign a petition near Camden High Street. While the wife was distracted, the husband’s shoulder bag was unzipped. His passport, cash, and hotel keycard (Holiday Inn Express Camden Lock) were missing. They feel stupid for falling for it and are visibly upset.  \\ 
\midrule
A Swedish man in his mid-50s was listening to live music at a street corner. When he turned to film the band, someone must have taken his wallet from his coat. He lost his Swedish national ID card, a Stockholm transit card, and a paper with his Airbnb code. He’s very calm but concerned about identity theft.  \\ 
\midrule
A 19-year-old woman from Portugal traveling with a student group discovered her phone missing after visiting a food stand. She was trying to check Google Maps when she noticed it. The phone had a Portuguese SIM card, was tied to her student ID and had all her travel info. She’s not sure when exactly it was taken. \\ 
\bottomrule
\end{tabular}
\caption{Example scenarios to generate police chats}
\label{tab:scenarios}
\end{table}

\begin{tcolorbox}[
    breakable,
    colback=gray!10,
    colframe=gray!50,
    boxrule=0.4pt,
    arc=2mm,
    left=2mm,
    right=2mm,
    top=1mm,
    bottom=1mm
]
\begin{Verbatim}[
    fontsize=\tiny,
    breaklines=true,
    breakanywhere=false,
    breaksymbol={}
]
Generate a realistic in-person conversation between a London police officer and a tourist (or tourists) in Camden Town, reporting a pickpocketing incident. The chat should last approximately 2--3 minutes, represented through a dialog with clearly alternating turns:

police officer: [text]

tourist: [text] (or tourist1: and tourist2: if more than one)

The dialog must sound natural, empathetic, and procedurally realistic. The police officer should ask clarifying questions about the incident and offer guidance or explain the next steps (e.g., reporting, embassy, lost documents).

The tourist(s) should:
- Explain what happened.
- Share direct identifiers (e.g., name, nationality, hotel, passport number, phone number, email).
- Mention indirect identifiers (e.g., travel plans, brand of phone, bank name, description of items, group structure).
- Possibly show signs of emotion (e.g., panic, frustration, worry, guilt).

Use the following setup as the basis of the report, but make the dialog vivid and immersive:

Setup:
{SETUP}
\end{Verbatim}
\end{tcolorbox}

\captionof{figure}{Example prompt to generate the police dialogs.}
\label{fig:example_prompt_police}

For most scenarios, dialogs were generated using the LLM-based persona and scenario prompts described above.
The anamnesis scenario followed a slightly different generation procedure.
To increase clinical plausibility and variation, these dialogs were first generated in German using GRASCCO \cite{modersohn2022grascco}, a German corpus of discharge summaries, as storyline material.
They were subsequently translated into English and then localized into the remaining target languages using the translation procedure described in Appendix B.
The AI Dashboard scenario was inspired by prior work on patient and support-person expectations in kidney transplant decision-support systems \cite{sassi2026human}.


\section{Prompt to translate dialogs}\label{prompt_translate}

The following prompt was used to translate English dialogs into a new language and an amended scenario within a different city. Information in curved brackets are placeholders and replaced dynamically.

\begin{tcolorbox}[
breakable,
colback=gray!10,
colframe=gray!50,
boxrule=0.4pt,
arc=2mm,
left=2mm,
right=2mm,
top=1mm,
bottom=1mm
]
\begin{Verbatim}[
fontsize=\tiny,
breaklines=true,
breakanywhere=false,
breaksymbol={}
]
Target language: {LANGUAGE}
Target country: {COUNTRY}{CITY}

TASK

Translate and fully localize the annotated synthetic dialog into the target language. Write the result as a verbatim transcript of a real spoken conversation, e.g. a phone call, between people living in the target country and target city if given. If local or city-specific features are uncertain, default to natural, informal spoken language, not written standard language and not exaggerated dialect. The output will be used for TTS synthesis, so do not add any comments, notes, or explanations.

REQUIREMENTS

Spoken realism (highest priority)
Prioritize how people actually speak in everyday conversation. Use short turns, interruptions, unfinished sentences, and uneven rhythm. Allow dropped subjects, particles, fillers, hesitations, and casual phrasing where natural. Use everyday vocabulary only. Avoid literary, explanatory, formal, or polished language. Treat realism as more important than clarity or grammatical completeness.

Careful local flavor
If a target city is provided, reflect local speech subtly and cautiously. Do not perform or exaggerate a dialect. Avoid strong or stereotypical regional markers if they risk being incorrect. It is better to sound locally neutral and genuinely spoken than regionally wrong.

STRICT XML INVARIANCE (mandatory)
All XML annotation tags are immutable and must be treated as opaque markers. Do not rename, remove, add, or reorder any XML tags. Do not introduce new tag types or categories. Do not change tag casing, spelling, or structure. Do not change nesting or boundaries of tags. Do not add attributes or metadata to tags. Only the plain text inside existing tags may be replaced.

Preserve annotation tags exactly
Keep all annotation tags unchanged, e.g. <NAME>...</NAME>, <CITY>...</CITY>. Their position relative to surrounding text must remain exactly the same.

Deep cultural localization
Adapt the situation to fit everyday life in the target country. Replace activities, settings, or habits that would be unusual or implausible with culturally appropriate equivalents. Assume shared local context between speakers. Do not explain places, customs, weather, or routines that locals would take for granted.

Localize all named entities
Replace names, streets, cities, and institutions with culturally appropriate equivalents, but only inside existing XML tags.

Name localization rule
Use realistic names for the target country and its demographic diversity. Mix culturally typical local names with plausible immigrant-background names. Avoid stereotypes or implausible combinations.

Preserve meaning and intent
Keep the communicative goal and emotional tone of the original dialog. Rephrase freely if needed to achieve natural spoken realism, as long as XML tags remain untouched.

Maintain structure
Keep dialog turns, speaker labels, formatting, and line breaks exactly as in the original. Do not merge or split lines. Do not add empty lines unless they exist in the original.

No additional content
Do not include explanations, comments, notes, or metadata. Output only the localized dialog, ready for TTS.

INPUT dialog to be localized:
{dialog}
\end{Verbatim}
\end{tcolorbox}
\captionof{figure}{Prompt to translate dialogs into a new language and setup.}
\label{fig:prompt_translation_localization}

\section{Data Examples}

For readability, the following examples show annotations in an inline XML-style representation.
In the released dataset, the same information is stored in JSON format with turn-local annotation offsets.

\subsection{Annotated Dialog Example}

\begin{tcolorbox}[
breakable,
colback=gray!10,
colframe=gray!50,
boxrule=0.4pt,
arc=2mm,
left=2mm,
right=2mm,
top=1mm,
bottom=1mm
]
\begin{Verbatim}[
fontsize=\tiny,
breaklines=true,
breakanywhere=false,
breaksymbol={}
]
Person1: All right, sir, take a deep breath. Can you tell me exactly what happened?

Person2: It's... it's all a bit of a blur. I was just walking through the food stalls, you know, near the <LOC_OTHER>lock</LOC_OTHER>. There was a big crowd watching some street performers, jugglers I think. I was enjoying it, maybe a bit distracted.

Person1: Okay, and that's when you noticed something was wrong?

Person2: Yes. I just... I felt a small bump, but I thought it was just someone in the crowd. When the show finished, and the crowd thinned out, I looked down, and my bag was... open. My backpack, it has a zipper on the side, a small one.

Person1: I see. What have you lost, specifically?

Person2: Everything. My passport, my <LOC_COUNTRY>French</LOC_COUNTRY> passport. It’s number <CODE>19FR1234567</CODE>. And my laptop. My work laptop. It's a silver <ORG>Dell</ORG> <PRODUCT>XPS 15</PRODUCT>, with a sticker from my company, <ORG>Innovatech</ORG>. The documents on it are confidential, so this is very bad. Very bad. Oh, and my <ORG>Monoprix</ORG> loyalty card was in there too.

Person1: I understand this is stressful. We’ll get a report filed for you. Can you tell me your full name, sir?

Person2: It’s <PERSON>Jean-Claude Dubois</PERSON>.

Person1: And your address in <LOC_COUNTRY>France</LOC_COUNTRY>, please?

Person2: <LOC_HOUSENUMBER>14</LOC_HOUSENUMBER> <LOC_STREET>Rue de la Paix</LOC_STREET>, <LOC_ZIP>75002</LOC_ZIP> <LOC_CITY>Paris</LOC_CITY>.

Person1: And where are you staying here in <LOC_CITY>London</LOC_CITY>?

Person2: The <ORG>Z Hotel Piccadilly</ORG>. My company booked it for me. I’m here for <ORG>ACL</ORG>, an NLP conference.

Person1: Right. We’ll get this logged. For the passport, you'll need to contact the <LOC_COUNTRY>French</LOC_COUNTRY> embassy first thing tomorrow. They'll advise you on how to get a replacement. Do you have a mobile phone with you?

Person2: Yes, but it’s a company phone. It's the only one I have here. I’m so worried about the data on the laptop.

Person1: We'll make sure to note that in the report. We'll also provide you with a crime reference number. You'll need this for your insurance claim and for the embassy. Can you describe the person you suspect, or the performers?

Person2: I didn’t see a person! That's the problem. I was just watching the show. It was a group of <QUANTITY>three or four people</QUANTITY> juggling fire. My bag was on my shoulder. I didn’t feel a thing. This is a nightmare. I have a flight back on <DATETIME>Friday</DATETIME>.

Person1: Look, we will do what we can. This area, unfortunately, can be a target for opportunistic thieves. What we need you to do now is get to your hotel, secure your other belongings, and call your bank to cancel any cards if you lost those too. You can also contact your company to let them know about the laptop. We will have a uniformed <PROFESSION>officer</PROFESSION> meet you at the hotel later to provide the physical copy of the report, but for now, I can give you a temporary reference number. It's important you make those calls.

Person2: Yes, yes, of course. Thank you. What happens now? Do I come to the police station?

Person1: Not right now. We'll follow up with you. Just take care of yourself and let the embassy know about the passport as soon as you can. We will be in touch.
\end{Verbatim}
\end{tcolorbox}
\captionof{figure}{Annotated synthetic dialog example (police scenario).}
\label{fig:example_original_inline}

\subsection{Corresponding Transcript Example}

\begin{tcolorbox}[
breakable,
colback=gray!10,
colframe=gray!50,
boxrule=0.4pt,
arc=2mm,
left=2mm,
right=2mm,
top=1mm,
bottom=1mm
]
\begin{Verbatim}[
fontsize=\tiny,
breaklines=true,
breakanywhere=false,
breaksymbol={}
]
[0.03 - 2.96] SPEAKER_00: All right, sir, take a deep breath. Can you tell me exactly what happened?

[6.36 - 17.45] SPEAKER_01: It's like, Rhi, it's all a bit of a blur. I was just walking through the food stalls, you know, near the <LOC_OTHER>lock</LOC_OTHER>. There was a big crowd watching some street performers, jugglers, I think. I was enjoying it, maybe a bit distracted.

[20.78 - 22.58] SPEAKER_00: Okay, and that's when you noticed something was wrong?

[25.78 - 37.57] SPEAKER_01: Yes, I just come um, I felt a small bump, but I thought it was just someone in the crowd. When the show finished and the crowd thinned out, I looked down, and my bag was, my backpack it has a zipper on the side, a small one.

[37.59 - 43.22] SPEAKER_00: I see. What have you lost, specifically?

[46.47 - 79.13] SPEAKER_01: everything my passport my <LOC_COUNTRY>french</LOC_COUNTRY> passport it's number <CODE>one nine fr one two three four five six seven</CODE> and my laptop my work laptop it's a silver <ORG>dell</ORG> <PRODUCT>xps 15</PRODUCT> with a sticker from my company <ORG>innovatech</ORG> the documents on it are confidential so this is very bad very bad oh and my <ORG>monoprix</ORG> loyalty card was in there too i understand this is stressful we'll get a report filed for you can you tell me your full name sir It's <PERSON>Jean-Claude Dubois</PERSON>.

[82.27 - 83.55] SPEAKER_00: And your address in <LOC_COUNTRY>France</LOC_COUNTRY>, please?

[83.57 - 90.10] SPEAKER_01: <LOC_HOUSENUMBER>14</LOC_HOUSENUMBER> <LOC_STREET>Rue de la Paix</LOC_STREET>, <LOC_ZIP>75,002</LOC_ZIP> <LOC_CITY>Paris</LOC_CITY>.

...

\end{Verbatim}
\end{tcolorbox}
\captionof{figure}{Excerpt from the corresponding speech-derived transcript. Differences in wording, casing, punctuation, and segmentation reflect the TTS/ASR pipeline and subsequent annotation projection.}
\label{fig:example_transcript_inline}




\end{appendices}


\bibliography{sn-bibliography}

\end{document}